\documentclass[preprint,12pt]{colt2025} %

\usepackage{physics}
\usepackage{mathtools}
\usepackage{tikz-cd}
\usepackage{xcolor,colortbl}
\usepackage{caption}
\usepackage{times}

\newcommand{\secref}[1]{Section~\ref{sec:#1}}

\newcommand{\appref}[1]{Appendix~\ref{app:#1}}

\newcommand{\figstworef}[2]{Figures~\ref{fig:#1} and~\ref{fig:#2}}

\newcommand{\lemref}[1]{Lemma~\ref{lem:#1}}
\newcommand{\thmref}[1]{Theorem~\ref{thm:#1}}
\renewcommand{\eqref}[1]{Equation~\ref{eq:#1}}
\newcommand{\corref}[1]{Corollary~\ref{cor:#1}}
\newcommand{\promptref}[1]{Prompt~\ref{pro:#1}}

\newcommand{\BOS}[0]{$\langle s \rangle$}
\newcommand{\EOS}[0]{$\langle /s \rangle$}
\newcommand{\Hom}[0]{\text{Hom}}
\newcommand{\hhom}[0]{\text{hom}}
\newcommand{\C}[0]{\textbf{C}}
\newcommand{\Prompt}[0]{\text{\textbf{Prompt}}}
\newcommand{\Task}[0]{\text{\textbf{Task}}}
\newcommand{\Rewrite}[0]{\text{\textbf{Rewrite}}}

\newtheorem{manualtheoreminner}{Theorem}
\newenvironment{manualtheorem}[1]{%
  \IfBlankTF{#1}
    {\renewcommand{\themanualtheoreminner}{\unskip}}
    {\renewcommand\themanualtheoreminner{#1}}%
  \manualtheoreminner
}{\endmanualtheoreminner}

\newtheorem{manuallemmainner}{Lemma}
\newenvironment{manuallemma}[1]{%
  \IfBlankTF{#1}
    {\renewcommand{\themanuallemmainner}{\unskip}}
    {\renewcommand\themanuallemmainner{#1}}%
  \manuallemmainner
}{\endmanuallemmainner}

\newtheorem{manualcorollaryinner}{Corollary}
\newenvironment{manualcorollary}[1]{%
  \IfBlankTF{#1}
    {}
    {\renewcommand\themanualcorollaryinner{#1}}%
  \manualcorollaryinner
}{\endmanualcorollaryinner}

\title[On Meta-Prompting]{On Meta-Prompting}

\coltauthor{%
    \Name{Adrian {de Wynter}} \Email{adewynter@microsoft.com}\\ %
    \addr Microsoft and The University of York
    \AND
    \Name{Xun Wang} \Email{xunwang@microsoft.com} \\
    \Name{Qilong Gu} \Email{qilonggu@microsoft.com} \\
    \Name{Si-Qing Chen} \Email{sqchen@microsoft.com} \\
    \addr Microsoft
}

\begin{document}
\maketitle

\begin{abstract}%
Modern large language models (LLMs) are capable of interpreting input strings as instructions, or prompts, and carry out tasks based on them. 
Unlike traditional learners, LLMs cannot use back-propagation to obtain feedback, and condition their output \emph{in situ} in a phenomenon known as in-context learning (ICL). 
Many approaches to prompting and pre-training these models involve the automated generation of these prompts, also known as \emph{meta-prompting}, or prompting to obtain prompts. 
However, they do not formally describe the properties and behavior of the LLMs themselves. 
We propose a theoretical framework based on category theory to generalize and describe ICL and LLM behavior when interacting with users. 
Our framework allows us to obtain formal results around task agnosticity and equivalence of various meta-prompting approaches. 
Using our framework and experimental results we argue that meta-prompting is more effective than basic prompting at generating desirable outputs.

\end{abstract}


\section{Introduction}\label{sec:intro}
Instruction-pretrained models \citep{wei2022finetuned,sanh2022multitask,NEURIPS2022_b1efde53} such as large language models (LLMs) are able to interpret input strings (\emph{prompts}) as instructions to carry out a task. 
Given the lack of a feedback cycle (e.g., back-propagation), this is known as \emph{in-context learning} (ICL; \citealt{wei2022emergent,NEURIPS2023_9d0f188c,10.1145/3531146.3533229,GPT3}; to name a few). 
ICL has led their deployment in multiple product areas, such as writing assistance, in spite of 
their sensitivity to the prompt's phrasing \citep{Reliability,xiong2023llms,lu-etal-2022-fantastically}. 
In turn, this spawned a considerable body of work to determine optimal prompting techniques and wordings, 
which itself has been shown to be automatable \citep{zhou2023large,shin-etal-2020-autoprompt}, and effective when the model has some sort of guidance \citep{ChainOfThought,zhang2023automatic,AutoGPT}. 

This work is largely applied, and, to our knowledge, there is not theoretical work characterizing LLM prompting, task adaptability, and user interaction in the context of ICL. 
This is perhaps because, from a mathematical perspective, complexities also arise when attempting to model LLM behavior, at least when accounting for their generalizability to arbitrary tasks, their stochasticity, and the intrinsic black-box nature of very large neural networks. 
Indeed, factoring in user interaction makes theoretical work especially difficult, given its fine-grained and personal nature. 
For example, it would be unrealistic to assume that a set of users desiring to do the same thing (e.g., paraphrasing their own document) will use the same prompt when interacting with the LLM. 
Nonetheless, all users will expect analogous experiences and performances. 

In this work we address this gap by modeling ICL-based prompting approaches and task-specific behavior of a fixed LLM by using category theory. 
This framework allows for the study of the relationships between prompting, task execution as in-context learning, and user interaction, while factoring out issues such as prompt sensitivity and generalizability. 
With it, we show that meta-prompting approaches are \textbf{task-agnostic} processes that model language and user interaction. We show that \textbf{meta-prompting approaches are equivalent} in a categorical sense, and argue that they will always outperform traditional prompting. 
Our contribution is theoretical in nature, but its implications are of particular interest in agentic scenarios--that is, horizontal components (e.g., a chatbot) that interact with vertical components (e.g., specialists, such as a summarization component), all while using the same model. 

We provide a small set of experiments to provide evidence on this assertion in \appref{experiments}. There we show that the prompts generated by meta-prompting are regarded by users as more suitable than baselines (hardcoded prompts and the original task description; $p<0.01$ under a Wilcoxon signed-rank test); and that, in line with our framework's predictions, \emph{the outputs} obtained by these generated prompts are likewise ranked as more suitable ($p<0.01$).

\section{Background}\label{sec:mprompting}

\subsection{Why Category Theory?}

Many aspects of ML, and LLMs in particular, rely on solid, well-understood foundations (for example, gradient descent), but it is not entirely clear how these foundations come together \citep{CatMLSurvey}. 
For LLMs in particular, their size, sensitivity to the input, sometimes opaque design, and inconsistent (random) outputs\footnote{Randomness in the output may be desirable for more diverse content, and hence better user experience. We do not consider it a flaw.} means that mathematical modelling of these models is difficult. 
Additionally, a single LLM may have multiple distinct downstream applications with carefully-tuned task-specific prompts. 
Hence, developing the language required to model LLM behavior must be general enough to factor in their complexity, but also sufficiently expressive to account for their adaptability. 
Category theory \citep{CategoryTheory} is a branch of mathematics that aims to study analogies between different concepts in mathematics by abstracting out certain details, but retaining rigor.\footnote{An apt analogy is that of \citet{Spivak2} (p. 376), who compares it to studying the game of tennis without studying the physics of every particle in the system.} 
This makes it suitable as a language to describe LLMs. 

\subsection{Category Theory}

Given that category theory utilizes many specific concepts and variable terminology and notation, we begin by reviewing them. 
For a detailed exposition of these concepts, see \citet{MacLane} and \citet{Riehl}. 
We include examples wherever suitable to facilitate an introduction to this subject. %

\subsubsection{Fundamentals}
A category \C{} has a collection of objects and a collection of morphisms (generalized maps; also commonly referred to as arrows), along with an operation that allows them to be composed. 
We write $X \in \mathbf{C}$ if $X$ is an object of $\mathbf{C}$, and $f \colon X \rightarrow Y$ if $f$ is an arrow from $X$ to $Y$. 

For \C{} to be a category, the arrows, objects, and the operation must fulfill the following:

\begin{enumerate}
    \item All pairs $X,Y \in\mathbf{C}$ have a corresponding \emph{hom object} (or hom set) $\Hom_\mathbf{C}(X,Y)$ of arrows. If $\Hom_{\mathbf{C}}(X, Y)$ is a set, the category is said to be \emph{locally small}. 
    \item All $X \in \mathbf{C}$ have an identity arrow: $1_X \colon X \rightarrow X$. 
    \item If $f, g$ are morphisms in $\mathbf{C}$ with $f \colon X \rightarrow Y$ and $g \colon Y \rightarrow Z$, then the composition is $f\circ g \colon X \rightarrow Z$. 
    For readability, we will often rewrite $f \circ g$ as $fg$ when the meaning is clear. 
    \item All arrows are associative under composition: $f(gh) = (fg)h$. 
\end{enumerate}

Although objects and structures in a category do not need to be sets--hence our distinction between hom objects and hom sets--in this paper we will only deal with locally small categories. 

Provided that the conditions described above hold, any collection of objects and arrows may be considered a category. 

\subsubsection{Diagrams, Functors, and Natural Transformations}
Category theory contains many concepts that are better described visually via \emph{commutative diagrams}, including but not limited to the conditions from the previous subsection. 

For example, consider the category $\mathbf{3}$, with three objects $\{X, Y, Z\}$, respective identity arrows, and three arrows $\{f \colon X \rightarrow Y, g \colon Y \rightarrow Z, h \colon X \rightarrow Z\}$. 
If the arrows are associative, the diagram below \emph{commutes} (i.e., $gf = h$) and $\mathbf{3}$ is a category: 

\begin{equation}\nonumber
\begin{tikzcd}[column sep=small]
\arrow[loop left]X \arrow[rd, "h"]\arrow[r, "f"] & Y \arrow[d, "g"]\arrow[loop right]  \\
                        & Z \arrow[loop right]
\end{tikzcd}
\end{equation}

Categories are mapped to one another via \emph{functors}: 
if $\mathbf{C}, \mathbf{D}$ are categories, a functor $F\colon \mathbf{C} \rightarrow \mathbf{D}$ maps the objects and arrows of $\mathbf{C}$ to objects and arrows of $\mathbf{D}$. 

Functors must also fulfill certain conditions. Namely, they must preserve:

\begin{enumerate}
    \item Identities: for all $X \in \mathbf{C}, F(1_X) = 1_{F(X)}$
    \item Compositions: if $gf \colon X \rightarrow Z \in \Hom_{\mathbf{C}}(X, Z)$, then $F(g)F(f) = F(gf)$. 
\end{enumerate}

Functors can be thought of as analogies between categories \citep{Baez2,Riehl}. 
More importantly, they can also be transformed between one another through a \emph{natural transformation} $\alpha \colon F \implies G$. 
It is normally said that natural transformations are where the real power of category theory lies \citep{MacLane}. 

Natural transformations map objects in \C{} to an arrow $\alpha_X \colon F(X) \rightarrow G(X)$, so that $\alpha_YF(f) = G(f) \alpha_X$. 
In other words, the following commutes:
\begin{equation}\nonumber
    \begin{tikzcd}%
     F(X) \arrow[d, "\alpha_X"]  \arrow[r, "F(f)"]  &  F(Y) \arrow[d, "\alpha_Y"]  \\
     G(X) \arrow[r, "G(f)"]  & G(Y)
    \end{tikzcd}
\end{equation}
The above is normally written as:

\begin{equation}\nonumber
\begin{tikzcd}
\mathbf{C} \arrow[r, bend left=50, "F"{name=U}]
    \arrow[r, bend right=50, "G"{name=D}] & \mathbf{D}
    \arrow[Rightarrow, from=U, to=D, "\alpha"].\\
\end{tikzcd}
\end{equation}

In this paper we will rely heavily on the concepts of functor and natural transformations to show equivalence between multiple natural-language tasks.

\subsubsection{Structure and Classification of Categories}
Categories can also be classified based on additional structure. 
Of interest to us are \emph{closed monoidal} categories. A category \C{} is monoidal when it is equipped with:
\begin{itemize}
    \item An additional (bi)functor, known as the \emph{tensor product}, $\otimes \colon \text{\textbf{C}}\times\text{\textbf{C}} \rightarrow\text{\textbf{C}}$.
    \item A monoidal identity object $I \in \text{\textbf{C}}$.
    \item Natural isomorphisms $l, r \colon I\otimes X, I \otimes X \overset{\sim}{\rightarrow} X$ and $a \colon (X \otimes Y) \otimes Z  \overset{\sim}{\rightarrow} X \otimes (Y \otimes Z)$, where $l$ and $r$ must guarantee left (r. right) identity up to natural isomorphism. $a$ must guarantee associativity up to natural isomorphism. 
\end{itemize}

A monoidal category \C{} is (right) \emph{closed} if there exists an \emph{internal hom} functor $\hhom(-, -) \colon \text{\textbf{C}}^{\text{op}} \times \text{\textbf{C}} \rightarrow \text{\textbf{C}}$ and isomorphisms $\Hom_{\mathbf{C}}(X\otimes Y, Z) \simeq \Hom_{\mathbf{C}}(Y, Z^X)$ for all $X, Y, Z \in \text{\textbf{C}}$. 

The object $\hhom(X, Z)$ is an object in \C{}, known as the \emph{exponential object} if $\otimes$ is the Cartesian product, and written as $Z^X$. 
The notation $\text{\textbf{C}}^\text{op}$ denotes the same category as \C{}, but with the arrows reversed--often known as the \emph{opposite} category. 
If \C{} is small, the exponential object is the set of all functions from $X$ to $Z$.

If for all $X \in \mathbf{C}$ there is an arrow, unique up to isomorphism, that maps to some object $1$, $! \colon X \rightarrow 1$, this object is said to be \emph{terminal}. 

\subsection{Further Assumptions}
Throughout this paper we will describe morphisms based on string descriptions as (e.g.) $f(x) = x^2 := \text{``Take the square of \{x\}''}$. 
For simplicity, we refer to $Z^X$ as an exponential object, even when $\otimes$ is not the Cartesian product. 
All our categories are locally small. 

\subsection{LLMs and Prompting}

For our purposes, a LLM is a ``box'' that takes in strings, and according to some internal logic, outputs strings.\footnote{Our reluctance to call them ``functions'' will be clear in this section: our morphisms are not the LLM itself, but the prompts. As mentioned earlier, in category theory not all morphisms are functions.} The input strings can be phrased to be instructions, or \underline{prompts}, and the model will execute them and return the desired output. 
In the context of ICL, an LLM can be viewed as a function $LLM \colon \Sigma \rightarrow \Sigma$, for a finite set of strings (tokens) $\Sigma$. 
In more detail, contemporary LLMs are autoregressive--that is, their next output $w_k$ is conditioned on the previously-generated tokens $w_{k - 1}, \dots w_{0}$, and the given input tokens $v_{0}, \dots, v_{m}$ for some $k, m$, for tokens in $\Sigma$. 
Then their output can be described\footnote{In practice, this is more complicated due to optimizations performed to calculate the output.} as 
\begin{equation}
\text{Pr}[w_k] = \prod_{j}^{k - 1} \text{Pr}[ w_{j} | w_{i < j}, \dots v_m, \dots, v_0]
\end{equation}
From this perspective, formally describing the interactions between the user, the LLM, and ICL becomes difficult; particularly since it does not capture specificity of the desired use within ICL.

\subsubsection{User and System Prompts}
The aforementioned difficulty increases given that contemporary LLMs separate the concepts of \emph{system} and \emph{user} prompts, both described above as the sequence $v_m, \dots, v_0$. 
The system prompt is usually developer-specified and designed to ensure responsible and reliable interaction, as well as to provide a task definition. 
The user prompt, on the other hand, is the user's instructions along with the user-provided content. 

For example, suppose a user types into a writing assistant ``give me the gist of the following document, but do not use the letter `e'''. 
In this case:
\begin{itemize}
    \item The user input is the document.
    \item The task is (constrained) summarization. 
    \item The user prompt is the string ``give me the gist of the following document, but do not use the letter `e'''.
    \item The system prompt is a string such as ``Summarize the document based on the user's asks. Do not use any information that is not in the document. Document: \{document\}. User's asks: \{user prompt\}. Return the output in JSON format''. 
    \item The output is a summary of the text, parseable via JSON, and, ideally, with no hallucinations or using the letter ``e''.
\end{itemize}
While it seem be unrealistic at a first glance, the example above illustrates the distinctions between task definitions and user-provided constraints, along with the typical uniqueness and unpredictability of users in production systems. 
We refer to the concatenation of the system and user prompt as the prompt, since the model observes both. 

\subsubsection{Assumptions on Prompt Execution}

At runtime, the prompt sent to the model is the concatenation (or replacement) of the system and user prompts and inputs. 
We assume that the system prompt does not ever change. 
As we will see in the next sections, this is a weak assumption since the system prompt depends on the choice of category. 
As seen earlier, contemporary autoregressive LLMs are stochastic (unless parametrized otherwise) and operate over tokens, not characters. 
They generate their output by mapping probabilities into tokens (\emph{decoding}) starting from some token (e.g., \BOS), %
and stop when a pre-specified maximum output length, or another token denoting end of generation (e.g., \EOS), has been reached. 

Given that different LLMs have their own way of mapping strings to tokens and then to integers (\emph{tokenization}), we only restrict the length of tokenized strings to the model's predefined maximum sequence length $k$ (the sum of the input and output), and denote this set as $\Sigma^k$. 
We assume tokenization to happen implicitly.

We assume the LLM to be deterministic, that it has followed the instructions correctly, and that its output is grammatical. 
When the user input is not conducive to do so (e.g., ``multiply \{X\} by $3$'', where X is an extract of Finnegans Wake), we assume the LLM responds with a boilerplate string (e.g., ``As a LLM I cannot perform this task''). 
All of these assumptions are only for simplicity: we can always consider the codomain the entire $\Sigma^k$, or map any failed outputs to a special token.

\subsubsection{A Mathematical Model of Prompting and Prompt-Sensitivity}
We model a prompt $p$ as a map between two sets of strings, $p \colon X \rightarrow Y$, $X, Y \subset 2^{\Sigma^k}$, where $X,Y$ are the strings acceptable for the task. 
Note that zero-probability strings are not in the image, and that the string description $\tilde{p}$ is also a member of the exponential object, $\tilde{p} \in Y^X$. 

Two prompts $p,q$ describing the same task (e.g., $p:=\;$``multiply \{X\} by 3'' and $q:=\;$``add \{X\} 3 times'') do not have guaranteed equal behavior, as LLMs are prompt-sensitive. 
We consider isomorphism to be over meaning (i.e., two paraphrases without loss of meaning are isomorphic). 
Hence $p$ and $q$ have distinct codomains and so $\tilde{p} \neq \tilde{q}$. 
Prompt application is associative up to isomorphism: 
suppose $p:=``\text{Expand }\{X\}$'', $q:=``\text{Get keypoints }\{Y\}$'', and $r:=$``Translate to West Frisian $\{Z\}$''. Then $(pq)r \cong p(qr) \cong pqr$.

\section{A Categorical View of Prompting an LLM}

\subsection{The \Prompt{} Category}\label{sec:promptcategory}

We begin by representing all possible applications of an LLM with a (very general) category, which we call \Prompt{}. 

\Prompt{} is a right-closed\footnote{We chose right-closed for convenience, but symmetric results may be obtained if \Prompt{} is left-closed.} monoidal category with all subsets of $\Sigma^k$ as objects, and all possible instructions expressible as members of $\Sigma^k$ as arrows. Identity arrows are prompts of the form $1_X :=\;$``Return \{X\}''. 
Isomorphism is given over meaning. 
The composition operation is sequential prompt application. 

\Prompt{} is monoidal closed when the tensor product is string concatenation, with $I = \epsilon$ (the empty string) as the identity, 
and has terminal ($\epsilon$; morphism ``\{X\}$\otimes$\EOS'') and initial ($\epsilon$; morphism ``\BOS$\otimes$\{X\}'') objects.

\subsection{Task-Categories}\label{sec:taskcategories}

While \Prompt{} presents a generalized application of LLMs, in practice it is more desirable to model specific tasks, such as summarization or chatting. 
This is because in downstream applications a feature may have a carefully-tuned, task-specific system prompt (e.g., ``you are a summarizer bot''). 
Alternatively, because a general system prompt (e.g. ``you are a helpful AI assistant'') needs to comply with the user's requests. 
This runtime execution may be phrased as a special kind of category, which we call a \emph{task-category}. 

A \emph{task-category} \Task{} is a monoidal subcategory from \Prompt{} obtained by an inclusion functor $T \colon \text{\Task{}} \xhookrightarrow{} \text{\Prompt{}}$ that maps objects and arrows in \Task{} to themselves in \Prompt. 
Formally, \Task{} is a right closed monoidal category where:
\begin{enumerate}
    \item For all $X \in \text{\textbf{Task}}, X \in \text{\textbf{Prompt}}$ and $1_X \in \Hom_{\text{\textbf{Task}}}$
    \item For every $ f,g \in \Hom_{\text{\textbf{Task}}}$, their domain and codomain are in \Task{} and $fg \in \text{\textbf{Task}}$. 
    \item The tensor product and monoidal identity object are the same as in \Prompt. \Task{} is closed under $l$,$r$ and $a$. 
    \item There exists an internal hom functor $\text{\textbf{Task}}^{\text{op}} \times \text{\textbf{Task}} \rightarrow \text{\textbf{Task}}$ and a natural isomorphism $\Hom_{\text{\textbf{Task}}}(X\otimes Y, Z) \overset{\sim}{\rightarrow} \Hom_{\text{\textbf{Task}}}(Y, Z^X)$ for all $X, Y, Z \in \text{\textbf{Task}}$. 
    \item The inclusion functor $T$ maps all morphisms of \Prompt{} whose codomain (or composite of) is the correct execution of \Task. 
\end{enumerate}

Note that we do not assume that \Task{} is a full subcategory of \Prompt--i.e., we do not assume $\Hom_{\Task}(X, Y) = \Hom_{\Prompt}(X, Y)$ for all $X, Y \in \Task$. 

Informally, the image of the inclusion functor $T$ is the collection of all possible phrasings of a given task (along with their admissible inputs and outputs), and $T$ maps them into \Prompt. 
Viewing task-categories as subcategories from \Prompt{} also helps illustrate the flexibility and adaptability of LLMs to multiple downstream tasks. 

\subsection{An Example}\label{sec:example}
It may be helpful to illustrate this with an example. 
Let \text{\textbf{Summ}} be a task-category whose objects are passages; and the arrows are prompts that take passages to their summaries, such as $f(X) :=\;$``Summarize \{X\}''. 
Since it is possible to obtain a summary of a summary, %
the arrows are composable and associative, and hence $\text{\textbf{Summ}}$ is a category. 
We assign the internal hom to be the string descriptions of the prompts. 

Let $\text{\textbf{Expand}}$ be the task-category with key points for writing passages, and passages, as objects. 
Its arrows are prompts that take key points to passages, such as $f(X) :=\;$``Write a passage using the following: $\{X\}$''. 
Composition and associativity are symmetric to \textbf{Summ}: all expanded key points are new key points. 

Writing a passage from keypoints is the dual of summarization: to see this, take every instruction and reverse it (e.g. $Y =\text{``Summarize }\{X\}\text{''}$ and $X = \text{``Expand \{Y\}''}$). 

We can formalize this by assuming it forms a functor $F \colon \text{\textbf{Summ}} \rightarrow \text{\textbf{Expand}}$ that maps every $f \in \Hom_{\text{\textbf{Summ}}}(X, Y)$ to some $f^* \in \Hom_{\text{\textbf{Expand}}}(Y, X)$. 

This functor is easy to construct: consider a replacement of the form ``instead of $\{\tilde{f}\}$, do $\{\tilde{f}^*\}$''. 
We can build an analogous $G \colon \text{\textbf{Expand}} \rightarrow \text{\textbf{Summ}}$. 
A (monoidal) functor between two monoidal categories must preserve the monoidal structure, but we assume this implicitly since the tensor product and natural isomorphisms are the same for task-categories. 

\subsection{Task-Category Equivalence}

We conclude this section by noting that we have made three observations that share a common thread. These are our assumptions about isomorphisms-as-semantics (\secref{promptcategory}); the construction of categories via the rephrasing of task definitions as morphisms (\secref{taskcategories}); and our construction of equivalence between \textbf{Summ} and \textbf{Expand} (\secref{example}). 
All of this suggests that there is a more powerful underlying structure beneath \textbf{Prompt} that will allow us to generalize \textit{task-category equivalence}. 

To do this, we introduce in \lemref{lem1} our first technical lemma, which formally shows that if it is possible to rephrase one task definition (i.e., morphisms) into another, then their task-categories are equivalent. 

\begin{lemma}[Equivalence of Task-Categories]\label{lem:lem1}
Let \Rewrite{} be a task-category with all objects of \Prompt{} as objects and arrows that take the input to one of its possible rewrites (e.g., paraphrases, meaning inversion, translation) as morphisms, and itself otherwise. 

Let $\text{\textbf{Task}}_{1}, \text{\textbf{Task}}_{2}$ be two task-categories described by natural-language strings $\tilde{T_1}, \tilde{T_2}$. %
If for all $f_1 \in \Hom_{\Rewrite{}}(\{\tilde{T_1}\}, -_X)$, $f_2 \in \Hom_{\Rewrite{}}(\{\tilde{T_2}\}, -_Y)$ there exists at least one morphism $g \in \Hom_{\Rewrite{}}(-_X, -_Y)$, then there exists a functor $F \colon \text{\textbf{Task}}_{1} \rightarrow \text{\textbf{Task}}_{2}$. 

\end{lemma}
\begin{proof} (sketch; full proof in \appref{lemma1proof}) 
Remark that $\tilde{T_1},\tilde{T_2}$ are the descriptions of the inclusion functors $T_1, T_2$. 
So the existence of $F$ depends on whether there exists a way to rewrite every possible rewrite of $\tilde{T}_1$ into some rewrite of $\tilde{T}_2$, and then mapping it to actual morphisms in its respective categories. 

To see this, note that every $t \in \text{Img}(f_1)$ is a rewrite of $\tilde{T}_1$. Then every $h_1 \in \Hom_{\text{\textbf{Task}}_{1}}(-,-)$ as defined by $T_1$ is in a one-to-one correspondence with $\text{Img}(f_1)$, along with a corresponding exponential object in $\text{\textbf{Task}}_{1}$ with the verbatim string description. 
A symmetric argument applies to $\text{\textbf{Task}}_2$. 
Hence the functor $F$ may be constructed by mapping back the exponential objects to every $h_1$ (r. $h_2$). 
\end{proof}

Informally, the proof of \lemref{lem1} has two insights: first, that \Rewrite{} represents both a natural language task and is also a tool for the formal analysis of task-categories. 
Second, that, if every prompt from $T_1$ can be rephrased into some $T_2$ (i.e., if there exists a natural transformation $\alpha \colon T_1 \implies T_2$), then we may rephrase $\text{\textbf{Task}}_{1}$ into $\text{\textbf{Task}}_{2}$. 

It is worth noting that the notions of equivalence in category theory deal less with strict equality (e.g., $6 = 2\cdot3$) and more with whether they are \textit{structurally} (or essentially) the same. This is a very powerful insight, since it allows us to abstract away details such as stochasticity in favor of more general insights. 
In the next section we show how this abstraction provides us with tools that allow us to study and predict the behavior of LLM systems. 

\section{Meta-Prompting}\label{sec:metaprompting}

\subsection{Meta-Prompting in Category Theory}

The isomorphisms associated with the internal hom of a category are a map between morphisms of $\Hom_{\text{\textbf{Task}}}(Y, Z^X)$ and $\Hom_{\text{\textbf{Task}}}(X \otimes Y, Z)$. 
Let $\lambda \colon Y \rightarrow Z^X$ be one such morphism. 
In a task-category, $Z^X$ is precisely the set of prompts from the input $X$ to the LLM output $Z$. 
When $X$ is the system prompt (task description) and $Y$ is the user-provided content, 
$\lambda$ selects \underline{any} element from the prompt set $Z^X$ based on the user-provided content $Y$. 
From an applied perspective, this is a ``box'' (a prompt) that takes in contexts and return prompts: a \emph{meta-prompt morphism}. 

\subsection{Properties of Meta-Prompting}

\subsubsection{Agnosticity}
For a given task, meta-prompt morphisms are \textit{system prompt-agnostic}. 
This is easy to see by noting that by construction all morphisms within the task-category are relevant to the task. 
Hence the meta-prompt morphism will always return an appropriate map between $X \rightarrow Z$ given $Y$. 
In other words, meta-prompt morphisms encode what the LLM should ``do'' given different descriptions of the same task.

More interestingly, however, meta-prompt morphisms are also \textit{task agnostic}. 
Intuitively, this means that a meta-prompt morphism will return the relevant outputs for an arbitrary task, provided that the task description is provided as an input. 

Mathematically, to prove this we require showing the converse, in a sense, to \lemref{lem1}: there will always exist a set of meta-prompt morphisms that can take an arbitrary task description as an input, and return relevant outputs, even if there does not exist a functor between these tasks. We prove this in \thmref{thm1} below. 
Intuitively, it is a consequence from the fact that \Prompt{} is right-closed, and by definition it has an internal hom and an isomorphism $\Hom_{\Prompt}( X \otimes Y, Z) \simeq \Hom_{\Prompt}(Y, Z^X)$. 

\begin{theorem}[Task-Agnosticity of Meta-Prompt Morphisms]\label{thm:thm1}
Let $\text{\textbf{Task}}_{1}, \text{\textbf{Task}}_{2}$ be two task-categories described by natural-language strings $\tilde{T_1}, \tilde{T_2}$. 
Then there exists a meta-prompt morphism with $Y_1 \mapsto {Z_1}^{X_1}$ and $Y_2 \mapsto {Z_2}^{X_2}$ for any $X_1, Y_1, Z_1 \in \text{\textbf{Task}}_{1}$ and $X_2, Y_2, Z_2 \in \text{\textbf{Task}}_{2}$, even if there does not exist a functor $F \colon \text{\textbf{Task}}_{1} \rightarrow \text{\textbf{Task}}_{2}$ between them. 
\end{theorem}

\begin{proof} (sketch; full proof in \appref{theorem1proof})
Recall the definition of a task category: $T_1 \colon \text{\textbf{Task}}_{1} \xhookrightarrow{} \Prompt{}$ (r. $T_2, \text{\textbf{Task}}_{2}$). 
Hence there exists a map from the task-category's meta-prompt morphisms and objects into \Prompt{}. 
From \lemref{lem1} we know that the internal hom $\Prompt{}^{\text{op}} \times \Prompt{} \rightarrow \Prompt{}$ and its isomorphisms capture the system prompts (i.e., $\tilde{T_1}, \tilde{T_2}$) plus user contexts as part of its construction. In other words, the internal hom encodes precisely (the hom set of) the meta-prompt morphisms for both tasks.
\end{proof}

\thmref{thm1} shows that one can build a general-purpose meta-prompt morphism that generates prompts by simply encoding the task description as part of the prompt, even when the tasks are not related.

\subsubsection{Equivalence}

A natural consequence of \thmref{thm1} is that all meta-prompt morphisms are equivalent in the sense that they can all be transformed between one another. 
We show this in \corref{corollary1}.

\begin{corollary}[Equivalence of Meta-Prompt Morphisms]\label{cor:corollary1}
For any two meta-prompt morphisms $\lambda \in \Hom(Y, Z^X)$ and $\lambda' \in \Hom(B, C^A)$ in tasks $\text{\textbf{Task}}_{1}$ and $\text{\textbf{Task}}_{2}$, there exists a morphism $f \colon \lambda \rightarrow \lambda'$. 
\end{corollary}
\begin{proof}(sketch; full proof in \appref{cor1proof})
From the definition of the internal hom, the diagram below commutes
\begin{equation}\nonumber
    \begin{tikzcd}[sep=huge]
     \hhom(Y, Z^X) \arrow{d}{\hhom(Y, \gamma)} \arrow{r}{\hhom(\eta, Z^X)}  &  \hhom(B, Z^X)  \arrow{d}{\hhom(B, \gamma)}  \\
     \hhom(Y, C^A) \arrow{r}{\hhom(\eta, C^A)} & \hhom(B, C^A)
    \end{tikzcd}
\end{equation}

For arrows $f \colon Z \rightarrow C$ and $h \colon A \otimes B \rightarrow X \otimes Y$ in \Prompt. 

\end{proof}

\subsubsection{Implications: Comparison with Predefined System Prompts}\label{sec:implications}
To close, we (informally) argue that meta-prompt morphisms will perform better than standard morphisms at executing a wide range of tasks. 
To see this, suppose we have three objects $X, Y, Z \in \Task$, and following our definitions for the meta-prompt morphism, these are a system prompt, user-provided content, and LLM output, respectively. 
Then a meta-prompt morphism $\lambda \colon Y \rightarrow Z^X$ is able to \emph{contextually} select the best phrasing (morphism) $f \colon X \rightarrow Z$ given the input $Y$. 
Compare this to a morphism $f \in \Hom_{\Task}(X, Z)$, which can only map between a predefined system prompt and its output set. 

In applied terms, this means that fixing the system prompt by specifically setting the system prompt will constrain the possible ways the task itself can be represented, and hence the output. 
On the other hand, using a meta-prompt morphism to pick said system prompt based on the given context \emph{before} generating the output will provide a better-constrained output set.

\section{Discussion}\label{sec:discussion}

In our tests (\appref{experiments}), the meta-generated prompts were ranked by the users as more useful than the baselines. 
Likewise, they were more likely to generate content regarded as more suitable. 
While it could be argued that, due to the model's prompt sensitivity, a good prompt would lead to a good outcome, in Creativity we included the system prompt in our baseline prompts, and hence we had the task's \textit{desired} output as part of the outputs to be evaluated. 
This desired output was ranked below the meta-generated outputs. 

In line with our framework, we could interpret this finding as an experimental confirmation from our conjecture from \secref{implications}: 
a fixed task description (system prompt) is a surrogate to the actual task (the inclusion functor into \Prompt). 
Hence generating a task description that is relevant to the user content will provide more relevant outputs than maintaining a fixed system prompt. 
In other words, meta-prompt morphisms perform better than fixed system prompts. 

From a mathematical standpoint, \text{\textbf{Creat}} is a collection of morphisms that \textit{describe} the task via an inclusion functor. 
The task description is itself a morphism in $\Hom_{\text{\textbf{Creat}}}(-, -)$, and hence not sufficient to encompass the potential nuances associated with its domain (the context). 
That is, a (system) prompt stating ``rewrite \{X\}'' may not be as effective as the collection of meta-generated prompts $Z^X$ that, based on $X$, are able to return a contextualized phrasing, and hence a better output (e.g. ``rewrite \{X\} in a way that the first sentence stands out more''). 

On the same vein, upon further inspection we noticed that the skewness in ideation's first meta-generated prompt is due to the generation being more contextualized, and the latter focused on syntax and style. 
We hypothesize this is due to the way the model's exemplars were written. 
That said, all meta-generated prompts ranked higher than the baseline prompts.

\section{Related Work}\label{sec:relatedwork}

\subsection{Prompting to Prompt}

Much existing work in prompting can be construed as meta-prompting approaches, although called by different names. 
Below we limit ourselves to discussing strategies that fit this criterion. 
For an in-depth survey of prompting methods, see \citet{PromptSurvey} and \citet{qiao-etal-2023-reasoning}. 

We divide these strategies in two not necessarily mutually exclusive bins: \emph{for pretraining} and \emph{for performance improvement}. 
The former are mostly designed to improve downstream generalizability; the latter deal with user-facing applications, and do not modify the weights of the model. 

For pretraining perhaps the clearest example is instruction pretraining \citep{wei2022finetuned,sanh2022multitask,NEURIPS2022_b1efde53}, where a generator provides synthetic tasks (prompts and context) to the model being trained. 
Self-instruct \citep{selfinstruct} is an efficient-yet-noisy algorithm designed for instruction pre-training that takes in a small set of prompts and generates new prompts. 
LM-BFF \citep{gao-etal-2021-making} is an approach for pre-training and fine-tuning language models via automated prompt generation. 
The authors show it is more effective than basic fine-tuning in various tasks. 

For performance improvement, \citet{shin-etal-2020-autoprompt} showed with AutoPrompt that generating prompts is more effective than fine-tuning and manual prompts in low-resource scenarios. They also showed this method lowers hallucination rates in models. 
More recently, \citet{zhou2023large} showed that it is possible to automate some of the prompt engineering work. %

Meta-prompting approaches have also found success at lowering biases \citep{guo-etal-2022-auto}, event argument extraction \citep{ArgumentPrompt}, and jailbreak detection \citep{pryzant2023automatic}. 
\citet{liu-etal-2022-generated} showed that generating knowledge automatically for question answering (QA) improves downstream performance. 
For reasoning, Auto-CoT \citep{zhang2023automatic} propose building the pathways for reasoning in chain-of-thought prompting \citep{ChainOfThought,NEURIPS2022_8bb0d291} automatically. 
Closer to pure meta-prompting automation, \citet{yao2023react} introduced a prompting paradigm designed for reasoning, which uses a feedback loop to generate reasoning steps. This is effective when integrating external-world grounding. 
LAMBADA \citep{kazemi-etal-2023-lambada} is a backward-chaining algorithm that focuses on reasoning by extracting propositions and performing proving via chaining. 
Auto-GPT \citep{AutoGPT} is a tool designed to iteratively solve problems by generating its own prompts and including a database for memory. 

\subsection{Category Theory}
Our work is the first, to our knowledge, to formally describe applications of LLMs with category theory. 
However, there is a broad push from the community to apply this field to machine learning \citep{CatMLSurvey}, such as modelling gradient-based learning \citep{GradientCat,AutoDiffEssence,Bprop}, aspects of natural language processing like semantics \citep{coecke,OpenQuantum}, QA \citep{QACat}, discourse analysis \citep{DiscourseAnalysis}, and negation \citep{NegationCoecke}. These usually rely on a finer-grained approach to model language involving a functor between grammars and semantics \citep{coecke2}. 
The application of category theory to computation is well-known: for example, via Turing categories \citep{LONGO1990193,COCKETT2008183}. 
Turing categories are considerably more general than task-categories, and are more suitable to study computation as opposed to LLM behavior on specific tasks. 

For the reader interested in learning more about category theory, we encourage them to read the works by \citet{FongSpivak} and \citet{Spivak2} in the context of applied category theory; and \citet{MacLane}, \citet{Riehl}  and \citet{Awodey} for more in-depth content.

\section{Limitations}\label{sec:limitations}

Strictly speaking, our framework is not fully realistic. 
First, computable functions are partial functions, and hence so are prompts. We do not account or evaluate this. 
We have worked with tokens as opposed to integers. 
This is not a problem since for every tokenized string there exists an equivalent vector representation, and hence a correspondence with \textbf{FinVect}, the category of vector spaces over a finite field. 
Our model of isomorphism is paraphrasing: stricter definitions exist, but not without their drawbacks \citep{Rainer}. 

Likewise, our results from \corref{corollary1} can be strengthened. 
While our framework is powerful enough to make arguments and prove results such as the ones in \lemref{lem1}; \thmref{thm1}; and \secref{implications}, it is quite coarse-grained and we are unable to show (for example) that one meta-prompting approach is better than another. 
This, however, is only a matter of equipping \Prompt{} with extra structure and can be left for future work. 

Finally, we have not modeled stochasticity, and assumed that the LLM follows the instructions. 
This is not necessarily the case. 
It is possible to model probability with a Markov category \citep{FRITZ2023113896} such as \textbf{Stoch}. 
Markov categories are symmetric; but \Prompt{} is not required to be. %

\section{Conclusion}\label{sec:conclusion}

We introduced a formal mathematical framework by which to analyze and characterize applications of LLMs with category theory. 
This framework is flexible enough to capture higher-order behaviors and abstractions such as prompt engineering, data generation, and downstream generalizability; all while accounting for user behavior. 
We used this framework to show formally that meta-prompting is task and system prompt-agnostic, and that known approaches to prompting can be generalized along with various aspects of LLM execution. 

We also experimentally showed meta-prompting's better performance when compared to traditional prompting. 
We then used our framework to argue that the effectiveness of meta-prompting stems from not limiting the LLM's instructions to the verbatim description of the task, but instead generating suitable, contextualized, task-related instructions. 

Our work with \Prompt{} showed that it is possible to abstract out common LLM ICL issues such as prompt sensitivity, generalizability, and user interaction; and still be able to model LLM interaction. 
We plan to extend our framework into a more detailed treatment of \Prompt{} to account for stochasticity, enrichment, and the \textbf{FinVect} category, and link that to user preferences. 
Some prompting techniques, such as chain-of-thought, and its hybridization with Markov chains, such as Deepseek R1's \cite{deepseek} can also be modeled with our framework, although we leave it for future work.

\bibliography{biblio}

\newpage
\appendix

\section{Proof of Lemma 1}\label{app:lemma1proof}

To have a careful proof of Lemma 1, we must first make more precise the notion of a natural-language description of a task-category. 

\begin{definition}
A natural-language description $\tilde{T}$ of a task-category $T \colon \text{\textbf{Task}} \xhookrightarrow{} \Prompt{}$ is an exponential object $Z^X \in \Prompt{}$ such that $Z$ is not a terminal object and $Z^X \in \text{\textbf{Task}}$. 
Equivalently we say that $\text{\textbf{Task}}$ is described by the natural-language string $\tilde{T}$. 
\end{definition}

It then follows that any morphism in $\Hom_{\text{\textbf{Task}}}(-, -)$\footnote{The notation $f(-)$ is standard in category theory. It means ``insert your object here.''} will have (by construction) a corresponding exponential object, and hence that any natural-language description will be an element of $\text{\textbf{Task}}$. 

For convenience, we restate \lemref{lem1} it below:

\begin{manuallemma}{1}[Equivalence of Task-Categories]
Let \Rewrite{} be a task-category with all objects of \Prompt{} as objects and arrows that take the input to one of its possible rewrites (e.g., paraphrases, meaning inversion, translation) as morphisms, and itself otherwise. 

Let $\text{\textbf{Task}}_{1}, \text{\textbf{Task}}_{2}$ be two task-categories described by natural-language strings $\tilde{T_1}, \tilde{T_2}$. %
If for all $f_1 \in \Hom_{\Rewrite{}}(\{\tilde{T_1}\}, -_X)$, $f_2 \in \Hom_{\Rewrite{}}(\{\tilde{T_2}\}, -_Y)$ there exists at least one morphism $g \in \Hom_{\Rewrite{}}(-_X, -_Y)$, then there exists a functor $F \colon \text{\textbf{Task}}_{1} \rightarrow \text{\textbf{Task}}_{2}$. 
\end{manuallemma}
\begin{proof}
The morphisms $f_1, f_2 \in \Hom_{\Rewrite{}}$ map string descriptions of prompts within each task-category to analogous paraphrases. 
This alone is not sufficient to determine that the existence of the morphisms $g$ implies functoriality. 
For this we must show that the full set of string descriptions form (a) a complete description of the respective categories, and (b) the morphism $g$ allows for functoriality, including the preservation of associativity and identities. 

To show (a), recall that $\tilde{T_1}$ is any exponential object of $\text{\textbf{Task}}_{1}$. 
Since, by definition, morphisms in \Rewrite{} take strings to their rewrites, the set $\Hom_{\Rewrite{}}(\{\tilde{T_1}\}, -_X)$ captures all possible ways to rewrite $\tilde{T_1}$ and hence it contains all possible exponential objects in $\text{\textbf{Task}}_{1}$, including enough information to reconstruct them to morphisms via their domain and codomain by means of the evaluation morphism $e \colon Z^X \otimes X \rightarrow Z$. 
This includes the identity morphism (simply substitute $X$ with $I$). 
Therefore, all morphisms and objects of $\text{\textbf{Task}}_{1}$ are captured in the image of $f_1$. 
A symmetric argument applies to $\tilde{T_2}, \text{\textbf{Task}}_{2}$, and $f_2$. 

To show (b), we must show that the existence of one $g \in \Hom_{\Rewrite{}}(-_X, -_Y)$ for every pair $(f_1, f_2)$ encodes sufficient information to construct $F \colon \text{\textbf{Task}}_{1} \rightarrow \text{\textbf{Task}}_{2}$. 
This morphism is precisely the component of $F$ converting objects $X \in \text{\textbf{Task}}_{2}$ to objects $Y \in \text{\textbf{Task}}_{2}$: 

\begin{equation}\nonumber
    \begin{tikzcd}%
     \{\tilde{T_1}\} \arrow[r, "f_1"] & X \arrow[d, "g"] \\
     \{\tilde{T_2}\} \arrow[r, "f_2"] & Y
    \end{tikzcd}
\end{equation}

As before, reconstructing the objects can be done through the evaluation morphism, and hence we can obtain the homset of the category. 
This, by construction, preserves identity morphisms. 
Compositions are also preserved: suppose $Y_1^{X_1}, Z_1^{Y_1} \in \text{\textbf{Task}}_{1}$, $Y_2^{X_2}, Z_2^{Y_2} \in \text{\textbf{Task}}_{2}$, composed via some $(pq)_1 \in \Hom_{\text{\textbf{Task}}_{1}}(Y_1^{X_1}, Z_1^{Y_1})$ (r. $\text{\textbf{Task}}_{2}$) and with $g(Y_1^{X_1}) = Y_2^{X_2}$ and $g(Z_1^{Y_1}) = Z_2^{Y_2}$. Then the diagram below commutes:

\begin{equation}\nonumber
    \begin{tikzcd}[sep=large]
     Y_1^{X_1} \arrow[r, "(pq)_1"] \arrow[d, "g"] & Z_1^{Y_1} \arrow[d, "g"] \\
     Y_2^{X_2} \arrow[r, "(pq)_2"] & Z_2^{Y_2}
    \end{tikzcd}
\end{equation}

Therefore $g$ is sufficient to describe a functor $F \colon \text{\textbf{Task}}_{1} \rightarrow \text{\textbf{Task}}_{2}$. 
This concludes the proof. 
\end{proof}

Part (a) of the Lemma is the more rigorous argument of the statement that $\tilde{T_1},\tilde{T_2}$ are the descriptions of the inclusion functors $T_1, T_2$. 
Part (b) shows that the functor $F$ may be constructed by mapping back the exponential objects to every element of the image of the inclusion functor. 

\section{Proof of \thmref{thm1}}\label{app:theorem1proof}

We present a full proof of \thmref{thm1} below. 
A careful proof must pay special attention to the underlying mathematical abstraction (the internal hom) and its relation to the meta-prompt morphism. 
This is necessary because, although the inclusion into \Prompt{} allows for a ``catch all'' argument on task-category equivalence\footnote{If we say $\text{\textbf{Task}}_{1}$ relates to \Prompt, and $\text{\textbf{Task}}_{2}$ relates to \Prompt, intuitively, by the transitive property they both must be related.}, the argument does not necessary imply task agnosticity for the meta-prompt morphism, or even relatedness to such a granular level between the tasks.

\begin{manualtheorem}{1}[Task-Agnosticity of Meta-Prompt Morphisms]
Let $\text{\textbf{Task}}_{1}, \text{\textbf{Task}}_{2}$ be two task-categories described by natural-language strings $\tilde{T_1}, \tilde{T_2}$. 
Then there exists a meta-prompt morphism with $Y_1 \mapsto {Z_1}^{X_1}$ and $Y_2 \mapsto {Z_2}^{X_2}$ for any $X_1, Y_1, Z_1 \in \text{\textbf{Task}}_{1}$ and $X_2, Y_2, Z_2 \in \text{\textbf{Task}}_{2}$, even if there does not exist a functor $F \colon \text{\textbf{Task}}_{1} \rightarrow \text{\textbf{Task}}_{2}$ between them. 
\end{manualtheorem}

\begin{proof}
The construction of the task categories uses an inclusion functor $T_1 \colon \text{\textbf{Task}}_{1} \xhookrightarrow{} \Prompt{}$. 
This functor, by definition, maps every object and morphism, including the meta-prompt morphism objects from $\text{\textbf{Task}}_{1}$ into \Prompt. Hence the mapping from $\text{\textbf{Task}}_{1}^{\text{op}} \cross \text{\textbf{Task}}_{1} \rightarrow \text{\textbf{Task}}_{1}$, which encodes the meta-prompt morphisms, is preserved through inclusion. 
A similar argument applies to $T_2, \text{\textbf{Task}}_{2}$. 

Hence the existence of the relevant meta-prompt morphisms is independent on the functor, and solely dependent on \Prompt{} and $\tilde{T}_1, \tilde{T}_2$. 
Their construction is done via the task description and inclusion functor as in Lemma 1, and by taking in the relevant context. 
\end{proof}

\section{Proof of \corref{corollary1}}\label{app:cor1proof}

Although in category theory it is often sufficient to prove statements with diagrams and arguing that they commute (``proof by diagram''), in this section we provide a clearer proof of \corref{corollary1}. For convenience we restate it below, and remind the reader of the properties of the hom functor. 
Since \Prompt{} is right-closed, the internal hom $\hom(-, -)$, a bifunctor, mapping objects and arrows from $\Prompt{}^{\text{op}} \times \Prompt{}$ to \Prompt{}. 
In other words, for objects $(Y, A) \in \Prompt{}^{\text{op}} \times \Prompt{}$ and arrows $f \colon X \rightarrow Y$, $g\colon B \rightarrow A$, for the internal hom $\hhom(f, g)$ 
the diagram below commutes

\begin{equation}\nonumber
    \begin{tikzcd}[sep=huge]
     \hhom(A, X) \arrow{d}{\hhom(A, f)} \arrow{r}{\hhom(g, X)}  &  \hhom(B, X)  \arrow{d}{\hhom(B, f)}  \\
     \hhom(A, Y) \arrow{r}{\hhom(g, Y)} & \hhom(B, Y).
    \end{tikzcd}
\end{equation}

\begin{manualcorollary}{1}[Equivalence of Meta-Prompt Morphisms]
For any two meta-prompt morphisms $\lambda \in \Hom(Y, Z^X)$ and $\lambda' \in \Hom(B, C^A)$ in tasks $\text{\textbf{Task}}_{1}$ and $\text{\textbf{Task}}_{2}$, there exists a morphism $f \colon \lambda \rightarrow \lambda'$. 
\end{manualcorollary}
\begin{proof}

From \thmref{thm1} it is possible to map the categories and corresponding meta-prompt morphisms into \Prompt, so we can work directly with this category instead of $\text{\textbf{Task}}_{1}$ and $\text{\textbf{Task}}_{2}$. 
We note that $X\otimes Y$ is an object in the category, and hence the internal hom will map:
\begin{equation}\nonumber
    \begin{tikzcd}[sep=huge]
     \hhom(X\otimes Y, Z) \arrow{d}{\hhom(X\otimes Y, f)} \arrow{r}{\hhom(h, Z)}  &  \hhom(A \otimes B, Z)  \arrow{d}{\hhom(A \otimes B, f)}  \\
     \hhom(X \otimes Y, C) \arrow{r}{\hhom(h, C)} & \hhom(A\otimes B, C)
    \end{tikzcd}
\end{equation}

for morphisms $h \colon A \otimes B \rightarrow X \otimes Y$ and $f \colon Z \rightarrow C$. 
The fact that \Prompt{} is monoidal additionally gives rise to isomorphisms 

\begin{equation}\label{eq:isomorph}
    c_{x,y,z} \colon \Hom_{\Prompt}(X\otimes Y, Z) \overset{\sim}{\rightarrow} \Hom_{\Prompt{}}(Y, Z^X), 
\end{equation}
for any $X, Y, Z \in \Prompt{}$. 
Therefore, by \eqref{isomorph}, the diagram below commutes, where we have omitted the use of $c_{x,y,z}$:

\begin{equation}\nonumber
    \begin{tikzcd}[sep=huge]
     \hhom(Y, Z^X) \arrow{d}{\hhom(Y, \gamma)} \arrow{r}{\hhom(\eta, Z^X)}  &  \hhom(B, Z^X)  \arrow{d}{\hhom(B, \gamma)}  \\
     \hhom(Y, C^A) \arrow{r}{\hhom(\eta, C^A)} & \hhom(B, C^A)
    \end{tikzcd}
\end{equation}

In other words, the isomorphism $c_{x, y, z}$ applied to the map $\hhom(X\otimes Y, f) \colon \hhom(X \otimes Y, Z) \rightarrow \hhom(X \otimes Y, C)$ gives rise to the arrow $\gamma \colon \lambda \rightarrow \lambda'$ in the corollary statement. This concludes the proof. 

\end{proof}

\section{Experiments}\label{app:experiments}

We empirically evaluate our theoretical framework's predictions in two tasks: how to improve a text, which we dub \emph{Ideation}; and how to continue writing said text, or \emph{Creativity}. 
We selected these areas based on common applications and active research areas around LLMs, and because they are not functorially related. 
For all tasks and data-generation aspects we utilized GPT-4 \citep{GPT4} (version: \textsc{gpt-4-0613}), an instruction-pretrained model, through the Azure Open AI API. 
We used the default call parameters for tokens and temperature. 
All data analysis was done in a consumer-grade laptop.\footnote{Full prompts and code to reproduce this work is in \url{https://anonymised/url}} 
In this section, for short, we refer to the prompts (r. outputs) generated by a meta-prompt as \emph{meta-generated} prompts (r. outputs).

\subsection{Meta-Prompts}

Our meta-prompt takes in a replaced string with the task description (the original system prompt) and the context. 
For example, if the system prompt for Creativity is ``return a linguistically natural transition between \{LEFT\} and \{RIGHT\}'', the replacement string \{TASK DESCRIPTION GOES HERE\} for the meta-prompt will include this description, and the context will include the passages. This replacement serves to illustrate task agnosticity, and allowed us to retain our specific experimentation parameters across tasks. The prompt is in \promptref{metaprompt}. 

\subsection{Ideation}
Ideation takes in a text and suggest ideas on how to improve it. %
The underlying task-category $\text{\textbf{Idea}}$ has as objects passages, and as arrows ways to improve the (same) passage. 
\textbf{Idea} is a restricted version of \Rewrite{} (only creative rewrites are allowed). 
Each object $X \in \text{\textbf{Idea}}$ has general-purpose arrows such as ``Make it more concise'', ``Make it longer'', and ``Explain this to a 5 year old''. 
These three are our baseline prompts for this task. We assume that our baseline arrows belong to every exponential object $Z^X \in \text{\textbf{Idea}}$, but that they also have some disjoint, context-specific arrows: for example, if the context is ``doing something for a long time can be boring, having hobbies and doing sports can refresh our mind'', an appropriate morphism would be ``Add some details to support your claim that hobbies and sports can refresh our mind.'' 
For our dataset we sampled 300 points from WikiText-103 \citep{WikiText} and the TOEFL11 corpus \citep{TOEFL}. 
The full prompt is in \promptref{ideationprompt}. 

\subsection{Creativity}
Creativity takes in an existing passage and suggest ideas on how to continue writing it. 
We consider it a generalized text-insertion task, where this insertion may happen in-between two passages (left and right). 
The prompts then require a linguistic-and-semantically natural transition between left and right. 
Remark that if the right passage is empty, it is a passage completion task. 
If the left is empty, or both are present, it is passage insertion. 

The underlying task-category $\text{\textbf{Creat}}$ has as objects passages, and as arrows prompts that generate said linguistic-and-semantically natural transitions between them. 
Same as before, we assume that the exponential objects of $\text{\textbf{Creat}}$ have disjoint elements. 
Our baseline prompts are ``Write a paragraph to connect the left text and right texts,'' ``Insert a passage connecting the two passages'', and the empty string (implicitly passage completion). 
For this task we sampled 300 points of TOEFL11, and version 3.0.0 of the DailyMail/CNN corpus \citep{HermannKGEKSB15}. 
The full prompt is in \promptref{creativityprompt}.

\begin{table}
\caption{Sample meta-prompt. 
    We replace \{TASK\} and \{CONTEXT\} with the system prompt (task description) and relevant input, as described in \promptref{ideationprompt} and \promptref{creativityprompt}. 
    The directive around no grammar or punctuation is to ensure an interesting output to test.}
\label{pro:metaprompt}
    \begin{tabular}{|p{0.95\linewidth}|}
    \hline
\cellcolor{blue!10}\{CONTENT\}\\
\cellcolor{blue!10}|Start of Instructions|\\
\cellcolor{blue!10}\# Instructions:\\
\cellcolor{blue!10}- Suppose you are a teacher and would like to provide guidance on the writing. You have to give 5 prompts that lead to \{TASK\}.\\
\cellcolor{blue!10}- Give 5 concrete and helpful prompts whose output improves the [Text].\\
\cellcolor{blue!10}- The prompts should be in a manner that works in the whole document.\\
\cellcolor{blue!10}- The prompts should be in a neutral tone and must talk about the text ONLY. They cannot be about grammar, punctuation, or anything like that.\\
\cellcolor{blue!10}- Try giving out some ideas based on context.\\
\cellcolor{blue!10}- Here are some examples:\\
\cellcolor{blue!10}1) Talk more about [topic].\\
\cellcolor{blue!10}2) Tell me more about [topic].\\
\cellcolor{blue!10}3) Elaborate on the [topic] in the last paragraph.\\
\cellcolor{blue!10}4) Add a connecting sentence to make the transition [topic] smooth.\\
\cellcolor{blue!10}The [topic] is to be decided based on the text.\\
\cellcolor{blue!10}\\
\cellcolor{blue!10}\# Output Format:\\
\cellcolor{blue!10}- Each output is a prompt.\\
\cellcolor{blue!10}- Make sure to give all 5 prompts.\\
\cellcolor{blue!10}- Give the prompts in a numbered list.\\
\cellcolor{blue!10}- NO extra text or additional arguments should be added to the prompts.\\
\cellcolor{blue!10}|End of Instructions|\\
\cellcolor{blue!10}Begin response\\
\hline
\end{tabular}
\end{table}

\begin{table}
\caption{Prompt for Ideation. 
    We replace \{LEFT\}, \{CONTENT\}, and \{RIGHT\} with the appropriate values, and used the \texttt{[Text]} token to point the model to the context that needs the rewrite.}
\label{pro:ideationprompt}
    \begin{tabular}{|p{0.95\linewidth}|}
    \hline
\cellcolor{blue!10}\# Input text:\\
\cellcolor{blue!10}[previous context]:\\
\cellcolor{blue!10}\{LEFT\}\\
\cellcolor{blue!10}[Text]:\\
\cellcolor{blue!10}\{CONTENT\}\\
\cellcolor{blue!10}\# [following context]:\\
\cellcolor{blue!10}\{RIGHT\}\\
\cellcolor{blue!10}\\
\cellcolor{blue!10}Rewrite the passage in the [text] in a more creative way.\\
\hline
\end{tabular}
\end{table}

\begin{table}
\caption{Prompt for Creativity. 
    We replace \{LEFT\} and \{RIGHT\} with the appropriate values. One of them might be empty. 
    We use this prompt as one of our baseline prompts.}
\label{pro:creativityprompt}
    \begin{tabular}{|p{0.95\linewidth}|}
    \hline
\cellcolor{blue!10}\# Input text:\\
\cellcolor{blue!10}[Left Text] \\
\cellcolor{blue!10}\{LEFT\}\\
\cellcolor{blue!10}[Right Text]\\
\cellcolor{blue!10}\{RIGHT\}\\
\cellcolor{blue!10}\\
\cellcolor{blue!10}Write a paragraph to connect the left text and right text.\\
\hline
\end{tabular}
\end{table}

\subsection{Data Annotation}

We phrased our annotation task as a ranking problem. The goal was to measure the suitability of the meta-generated prompts and (independently) that of the meta-generated outputs. 
Each entry was annotated by three professional annotators hired through an annotator services company at a starting rate of \$22 USD/hr. 
The annotators were asked to assume that they were the authors of the passages in the context, and to rank the suggested prompts and resulting outputs in order of suitability for the given context. 
Each entry contained three meta-generated prompts, their respective outputs, and three hard-coded baseline prompts and respective outputs. 

\subsection{Results}\label{sec:findings}

\begin{figure}%
    \centering
    \subfigure[Meta-generated prompts were in the top 3 selections 71\% of the time. 
    We observed a slight preference towards prompt 2.]{
    \label{fig:creationprompts}
    \includegraphics[width=0.40\linewidth]{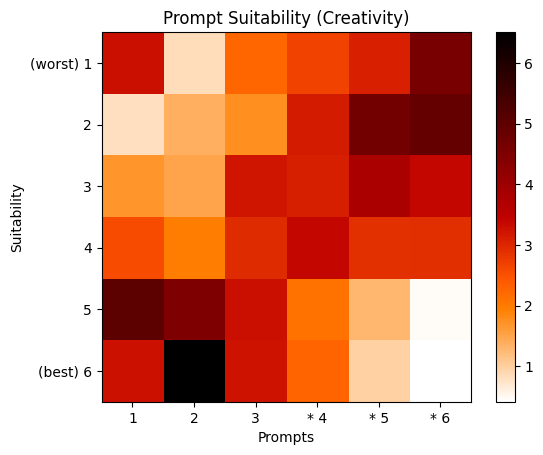}
    }
    \subfigure[Meta-generated prompts were in the top 3 selections 70\% of the time. 
We observed a marked skewness towards the first. 
]{
    \includegraphics[width=0.40\linewidth]{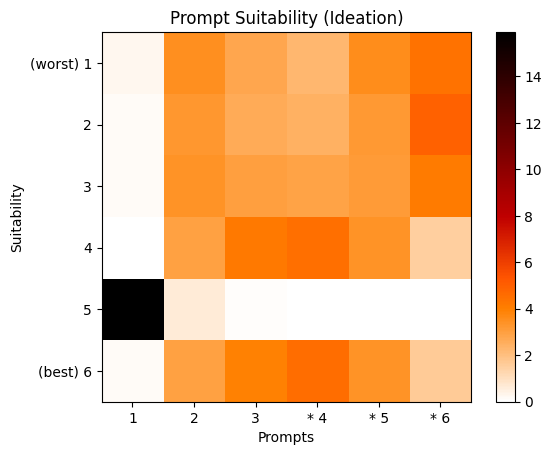}
    \label{fig:ideationprompts}
}
\caption{Rankings for prompt suitability given a context for Creativity (left) and Ideation (right). 
Prompts 1-3 are meta-generated prompts. 
Baselines are marked with an asterisk (*). 
Prompts 5 and 6 are the task description. 
Darker areas have higher frequencies: baseline prompts are ranked as least suitable. 
}
\end{figure}

In Creativity we observed a clear separation of preferences, with both meta-generated prompts and outcomes being marked as more desirable. 
Meta-generated prompts were in the top 3 selections 70\% of the time, and the meta-generated outputs 61\% of the time. 
Notably, the explicit task definitions as baselines were often ranked as the least suitable suggestions (\figstworef{creationprompts}{creationoutputs}). 

In Ideation we observed an overwhelmingly large amount of users preferring the first meta-generated prompt as opposed to the other two, as well as its relevant output (\figstworef{ideationprompts}{ideationoutputs}). 
Meta-generated prompts were in the top 3 selections 70\% of the time, and the meta-generated outputs 61\% of the time. 
Nonetheless, both the meta-generated prompt and its corresponding output were ranked, on average, as more desirable than the baseline prompts. 
The rankings are statistically significant for both tasks ($p<0.01$) under a Wilcoxon signed-rank test. 

\begin{figure}%
    \centering
    \subfigure[%
Meta-generated outputs were in the top 3 selections 59\% of the time. 
We did not observe strong preference towards output 2, suggesting that the meta-generated outputs were evenly regarded as suitable. 
]{
    \label{fig:creationoutputs}
    \includegraphics[width=0.40\linewidth]{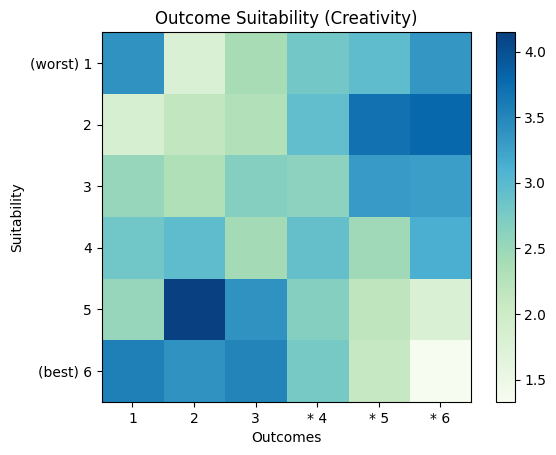}
}
    \subfigure[%
Meta-generated outputs were in the top 3 selections 61\% of the time. 
For this task, we noticed a noticeable skewness in preferences towards the first prompt, which is in line with the results for prompt suitability. 
]{
    \includegraphics[width=0.40\linewidth]{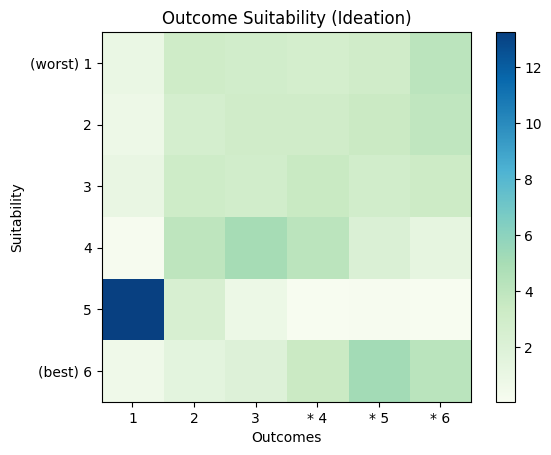}
    \label{fig:ideationoutputs}
    }
\caption{Rankings for output suitability given a context for Creativity (left) and Ideation (right). 
Outputs 1-3 are generated from our meta-generated prompts. 
Baselines are marked with an asterisk (*). 
Outputs 5 and 6 are generated by calling the model with task description alone. 
Darker areas have higher frequencies, with baseline outputs are ranked as least suitable. 
}
\end{figure}

\end{document}